# Optimal Design of a Walking Robot: Analytical, Numerical, and Machine Learning Methods for Multicriteria Synthesis

Arman Ibrayeva
*Institute of Mechanics and Engineering*
Almaty, Kazakhstan
sayatqyzy@gmail.com

Batyrkhan Omarov
*Institute of Mechanics and Engineering*
Almaty, Kazakhstan
batyahan@gmail.com

*Abstract*—This paper addresses several critical stages of designing a walking robot (WR), including optimal structural synthesis, introducing a novel "rational" mechanical structure aimed at enhancing efficiency and simplifying control system, while addressing practical limitations observed in existing designs. The study includes development of novel multicriteria synthesis methods for achieving optimal leg design, integrating analytical and numerical methods. In addition, a method based on Non-dominated Sorting Genetic Algorithm II is presented. Turning modes are investigated, and for the first time, the isotropy criterion, typically applied to parallel manipulators, is used for optimizing WR parameters to ensure optimal force and motion transfer in all directions. Several physical prototypes are developed to experimentally validate the functionality of different mechanisms of the robot, including adaptation to the surface irregularities and navigation using Lidar.

## I. Introduction

Leading groups in the field of research and development of WR include references [1 – 9]. Traditional approaches rely on active motor coordination, necessitating complex hierarchical control systems [10, 11]. Additionally, motors in such systems operate in intense acceleration and deceleration modes, resulting in *unreasonably low efficiency* [12]. The structural irrationality of most WRs is also linked to structural redundancy and multiple static indeterminacies. The number of active drives in biomorphic systems far exceeds the system's DoFs. For instance, anthropomorphic WRs, striving to replicate human bipedal walking and running, employ dynamic gaits and are highly expensive *despite achieving remarkable results* (e.g., Atlas). Similar challenges apply to quadrupedal and hexapedal designes characterized by the complexity of multi-level control systems, high energy consumption to support body weight. These factors collectively diminish the practicality of these systems.

An alternative approach is based on functional decomposition. Section II outlines the design principle of the WR based on rational structural synthesis, aimed at minimizing energy consumption and simplifying control. Structural synthesis addresses the issue of redundant connections and optimizes the number of motors required. An equivalent kinematic scheme (KS) of the robot is presented, alongside a model for a "safe" and efficient turning mechanism. Section III introduces developed methods and algorithms for structural-parametric synthesis across multiple criteria, that does not suffer from the common "branching defect" in existing methods, and presents the optimal design of the leg mechanism achieved. In Section IV, ML is applied to leg synthesis and compared with the aforementioned method. Section V investigates WR turning modes and develops a method for synthesizing WR parameters based on the isotropy criterion, defining optimal robot configurations for force transmission and motion efficiency. Section VI focuses on the development of a WR prototype with adaptability to terrain unevenness, with SLAM tested on a physical wheeled mobile robot, intended for future adaptation to the WR. This paper consolidates and enhances results previously published in [12 – 20], presenting integrated findings that collectively advance each discussed area.

## II. Rational KS of the WR

In traditional designs (universal-type WR), all motors are typically engaged simultaneously for both linear translational movement, rotational maneuvering, and adapting to uneven terrain. Standard structures like quadrupeds and hexapods commonly employ three to four motors per leg, resulting in a total of up to 24 motors. This section introduces a novel structure featuring only three primary motors for controlling horizontal movement and three *independent* adaptation motors, which operate without requiring coordination. This design reduces energy consumption by *eliminating motor reversal modes*, ensuring its stable motion instead, while also *eliminating structural redundancy that causes parasitic loads, deformation of links, and slippage of the robot's feet*. The subsequent sections focus on enhancing robot efficiency through further optimization of WR mechanism parameters.

In Fig. 1a, depicted is a first prototype, capable for rectilinear motion. As the crank's rotation angle $\varphi_{AB}$ varies within the limits of $\varphi_0 \leq \varphi_{AB} \leq \varphi_0 + \Phi_{sup}$, $\Phi_{sup} > \pi$, where $\Phi_{sup}$, defining the "duration" of the support phase (SPh), the connecting rod *EM* undergoes rectilinear translational motion. Consequently, all points on this linkage trace identical trajectories. By adding a pair *M*, adaptability to surface unevenness is ensured without compromising the rectilinearity of the body's motion (Fig.1b). During strictly vertical lifting and lowering of the foot, the absolute horizontal velocity of the foot *S* remains zero.

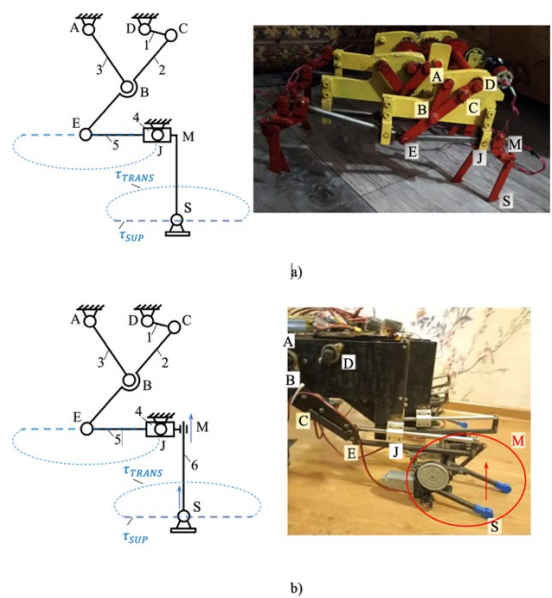

Fig. 1. Model of a WR using straight-line guide mechanism as a propulsor

For rotation of the body, we introduce additional joints $O_i$, with vertical axis $O_i z'_i$ (Fig.2a). To simplify the study of the turning modes, an equivalent KS of the WR is also presented (Fig.2b, 2c), where entire support-locomotion mechanisms (SLM) of the robot were modeled (depicted) as prismatic pairs $P_i$ ($i = 1, ..., 6$). Actuating the joints $O_i$ to turn the robot will again lead to structural redundancy. Thus, turning is carried out *due to the difference in velocities* of $P_i$.

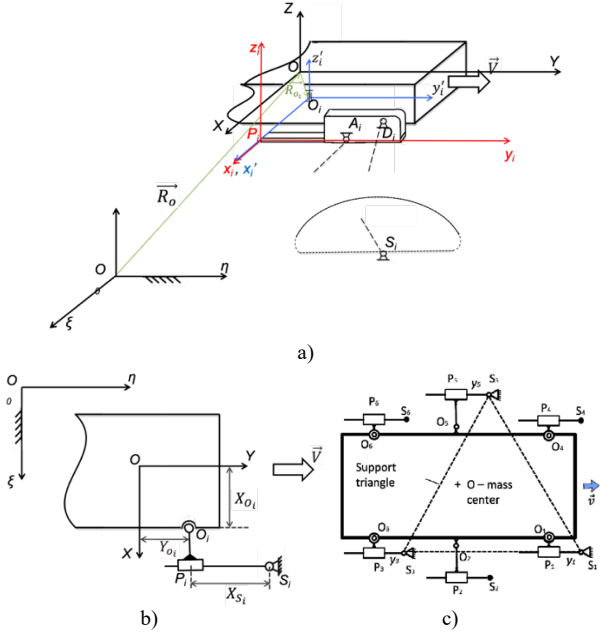

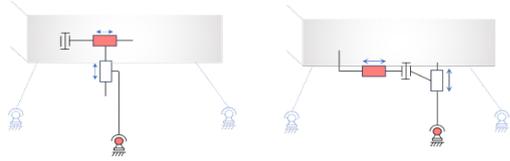

Fig. 2. An equivalent KS of the WR structural scheme on the plane $O\xi\eta$

Fig.3 illustrates an equivalent kinematic scheme in 3D space. *Only supporting legs* are depicted in the Fig.3, as legs in the transference phase do not contribute to the overall motion of the robot.

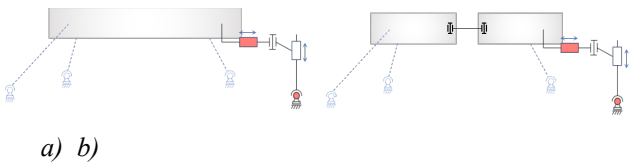

Fig. 3. An equivalent spacial KS of the WR, featuring only the supporting legs during "tripod" gate

The mechanism in Fig.2c has $W = 3$ DoF, since the number of moving links is $n = 7$, and $p_5 = 9$ ($p_5$ is the number of kinematic pairs with five constraints):

$$W = 3n - 2p_5 = 21 - 18 = 3,$$

For the spatial scheme:

$p_5 = 3\text{(legs)} \cdot 3\text{(per leg)}; W = 6(n+1) - 3p_3 - 5p_5 = 6(3 \cdot 3 + 1) - 3 \cdot 3 - 5 \cdot 9 = 6 = $ number of input joints.

In the case of 8-legged robot, the body must be segmented to meet the criterion of "rationality".

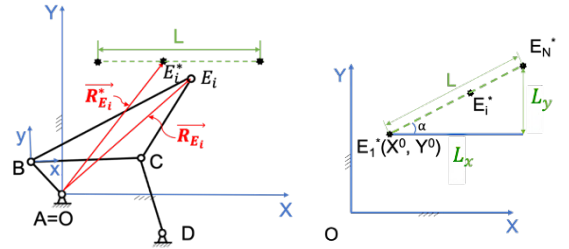

*a)  b)*

Fig. 4. An equivalent KS of the 8-legged WR during regular 4-legged gait

The following structural formula proves the "irrationality" of the structure shown in Fig.4a:

$p_5 = 4$ (supporting legs) $\cdot 3$;

$W = 6(n+1) - 3p_3 - 5p_5 = 6(4 \cdot 3 + 1) - 3 \cdot 4 - 5 \cdot 12 = 6 \neq$ number of input links $=12$

After partition of the robot body (Fig.4b):

$p_5 = 4$ (supporting legs) $\cdot 3 + 2; W = 6(n+1) - 3p_3 - 5p_5 = 6 \cdot 15 - 3 \cdot 4 - 5 \cdot 14 = 8 = $ number of input links.

Thus, the following decomposition of the robot's movement is proposed:

- generating rectilinear motion of the robot using a trajectory SLM;
- introducing additional drives to adapt to uneven terrain. These drives operate *independently* and *coordination of their work is not required*;
- controlling turns due to the difference in the angular velocities of the SLM cranks, without additional motors.

### III. DEVELOPMENT OF ANALYSIS AND SYNTHESIS METHODS FOR OPTIMAL LEG DESIGN

Consider a linkage *ABCD* with the linkage dimensions $X_A = Y_A = 0, X_D, Y_D, r_{AB}, l_{BC}, l_{CD}$ (Fig.5), where $X_A, Y_A, X_D, Y_D$ are the absolute coordinates of the frame pivots $A$ and $D$. The crank angle $\varphi_{AB}$ is varied within the range $\varphi_0 \leq \varphi_{AB} \leq \varphi_0 + \Phi_{\text{SUP}}$ as

$$\varphi_i = \varphi_0 + \Phi_{\text{SUP}} k_i, \text{where } k_i = \frac{i-1}{N-1}, i = 1, 2, ..., N.$$

and for each crank angle $\varphi_{AB} = \varphi_i$ position analysis of the mechanism is supposed to be carried out;

Fig. 5. KS of the linkage *ABCD* and the output trajectory in general case

In general case, if the trajectory is a straight line at angle $\alpha$ with horizontal and with length $L$, then

$$\begin{cases} X^*_{E_i} = X^0 + L_X k_i, i = 1, 2, ..., N \\ Y^*_{E_i} = Y^0 + L_Y k_i \end{cases} \quad (1)$$

where $L_X = L \cos \alpha$, $L_Y = L \sin \alpha$ (Fig. 5).

The absolute coordinates of the points $E_i(X_{E_i}, Y_{E_i})$ are

$$\vec{R}_{E_i} = \vec{R}_{B_i} + \Gamma(\beta_i)\vec{r}_E, \quad (2)$$

where $\vec{R}_{B_i}(X_{B_i}, Y_{B_i})$ is a radius vector of the joint $B$. $\Gamma(\beta_i)$ is 2×2 rotation matrix, $\vec{r}_{E_i}$ is a radius-vector of the coupler point ($E$) on real trajectory relative to $Bxy$. The synthesis task consists of satisfying approximately the following conditions:

$$\vec{R}_{E_i} - \vec{R}^*_{E_i} = \vec{0} \quad (3)$$

Consider a normalized mechanism, where $AD=1$. Varying parameters are: $x_1 = x_E, x_2 = y_E, x_3 = X^0, x_4 = Y^0, x_5 = L_x, x_6 = L_y$ (Fig.5). Then based on (3) *the initial* minimized approximation function is

$$\sum_{i=1}^{N} \left\{ (X_{B_i} + x_1 \cos\beta_i - x_2 \sin\beta_i - x_3 - x_5 k_i)^2 + (Y_{B_i} + x_1 \sin\beta_i + x_2 \cos\beta_i - x_4 - x_6 k_i)^2 \right\} \quad (4)$$

The necessary minimum condition $\partial \delta / \partial x_j = 0, j = 1, \ldots, 4$, yield to the linear equations in the form:

$$AX = b, \quad (5)$$

$$\mathbf{A} = \begin{bmatrix} \mathbf{E} & \mathbf{A}_1 & \mathbf{A}_2 \\ \mathbf{A}_1^T & \mathbf{E} & 1/2\mathbf{E} \\ \mathbf{A}_2^T & 1/2\mathbf{E} & \frac{1}{N}\sum_{i=1}^{N} k_i^2 \mathbf{E} \end{bmatrix}, \mathbf{E} = \begin{bmatrix} 1 & 0 \\ 0 & 1 \end{bmatrix},$$

$$\mathbf{A}_1 = \begin{bmatrix} -\frac{1}{N}\sum_{i=1}^{N} \cos\beta_i & -\frac{1}{N}\sum_{i=1}^{N} \sin\beta_i \\ \frac{1}{N}\sum_{i=1}^{N} \sin\beta_i & -\frac{1}{N}\sum_{i=1}^{N} \cos\beta_i \end{bmatrix},$$

$$\mathbf{A}_2 = \begin{bmatrix} -\frac{1}{N}\sum_{i=1}^{N} k_i \cos\beta_i & -\frac{1}{N}\sum_{i=1}^{N} k_i \sin\beta_i \\ \frac{1}{N}\sum_{i=1}^{N} k_i \sin\beta_i & -\frac{1}{N}\sum_{i=1}^{N} k_i \cos\beta_i \end{bmatrix} \quad (6)$$

$$\mathbf{b} = [b_1, b_2, \ldots, b_6]^T,$$

$b_1 = -\frac{1}{N}\sum_{i=1}^{N}(X_{B_i}\cos\beta_i + Y_{B_i}\sin\beta_i),$

$b_2 = \frac{1}{N}\sum_{i=1}^{N}(X_{B_i}\sin\beta_i - Y_{B_i}\cos\beta_i),$

$b_3 = \frac{1}{N}\sum_{i=1}^{N} X_{B_i}, b_4 = \frac{1}{N}\sum_{i=1}^{N} Y_{B_i},$

$b_5 = \frac{1}{N}\sum_{i=1}^{N} k_i X_{B_i}, b_6 = \frac{1}{N}\sum_{i=1}^{N} k_i Y_{B_i}$

However, solving Equation (6) does not provide an exact generation of the desired foot path, but supplies an approximate solution. In addition, achieving the best accuracy conflicts with some additional design requirements, such as force transmission, step height, stability, etc. We can address multiple conflicting criteria and improve accuracy taking into account that the matrix *A* and the vector *b* depend on five non-linear parameters: $p_1 = \frac{l_{AB}}{l_{AD}}, p_2 = \frac{l_{BC}}{l_{AD}}, p_3 = \frac{l_{CD}}{l_{AD}}, p_4 = \varphi_0, p_5 = \Phi_{SUP}$. Then the modified objective function is

$$\delta^0 = \min_{\vec{x}} \delta(\vec{P}, \vec{x}(\vec{P}))$$

$$\delta^0 = \delta\left(\vec{P}, \vec{x}(\vec{P})\right) = \delta\left(\vec{P}, A^{-1}(\vec{P}) \cdot b(\vec{P})\right) \to min.$$

By integrating the described analytical method with numerical, based on Sobol-Statnokov's LP-tau sequences, we synthesized the mechanism with desired accuracy of the output motion, transmission angle (increased from $12° - 15°$ in existing prototypes to $25°$) and step cycle parameter. The step cycle parameter $\nu = \Phi_{SUP}/\Phi_{TRANSF}$ of the synthesized mechanism is 1.59, the duration of the support phase is increased up to $221°$, whereas the values of these parameters in prototype are $\nu = 1.045$ and $\Phi_{SUP} = 184°$.

## IV. NSGA-II FOR LEG SYNTHESIS

The design of the leg mechanisms for WR is an optimization problem that requires the achievement of a set of conflicting objectives; for instance, propulsion, stability, and efficiency. About solving such multi-objective optimization problems a well developed framework is offered by the Non-dominated Sorting Genetic Algorithm II (NSGA-II). This section provides more details on the use of NSGA-II to synthesize leg linkages and demonstrate the usage of the algorithm's ability to identify Pareto optimal solutions that meet the needs of particular design objectives.

### A. Mathematical Formulation

The design problem can be formulated as a multi-objective optimization problem, expressed as in equation (7):

$$\min F(x) = [f_1(x), f_2(x), \ldots, f_m(x)] \quad (7)$$

Subject to

$$g(x) \leq 0, \quad h(x) = 0, \quad x^{(L)} \leq x \leq x^{(U)} \quad (8)$$

where *x* is the vector of design variables, $f_i(x)$ are objective functions to be minimized, $g(x)$ and $h(x)$ are inequality and equality constraints, $x^{(L)}$ and $x^{(U)}$ are the lower and upper bounds of the design variables.

For leg linkage design, the following objectives are the main goals: average the foot point trajectory approximation error ($\epsilon$) to the least possible gradient; maximize the force transmission angle ($\mu$). The approximation error is defined as

$$\varepsilon = \frac{1}{N} \sum_{i=1}^{N} \|p_i - \hat{p}_i\|^2 \quad (9)$$

Where $p_i$ are the desired positions, and $\hat{p}_i$ are the actual positions of the foot center.

The force transmission angle $\mu$ is given by:

$$\mu = \arctan\left(\frac{F_{vertical}}{F_{horizontal}}\right) \quad (10)$$

where $F_{vertical}$ and $F_{horizontal}$ are the vertical and horizontal components of the force exerted by the leg mechanism.

### B. NSGA-II Implementation and the Results

Fig.6 demonstrates the block scheme of the proposed NSGA-II solution. In Fig.7, an empirical correlation is manifested between authentic data, as procured from the sampling table, and solutions generated employing the Non-dominated Sorting Genetic Algorithm (NSGA-II) method.

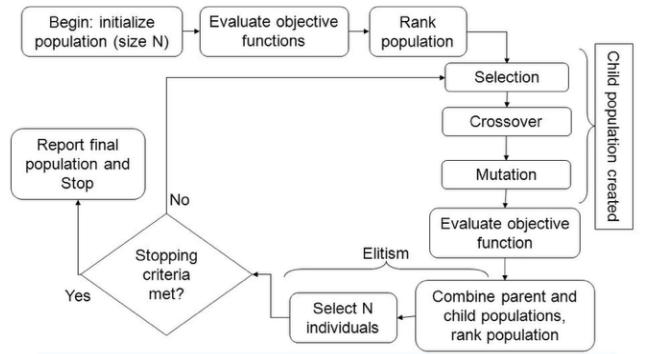

Fig. 6. Flowchart of non-dominated sorting genetic algorithm for Pareto optimal solution

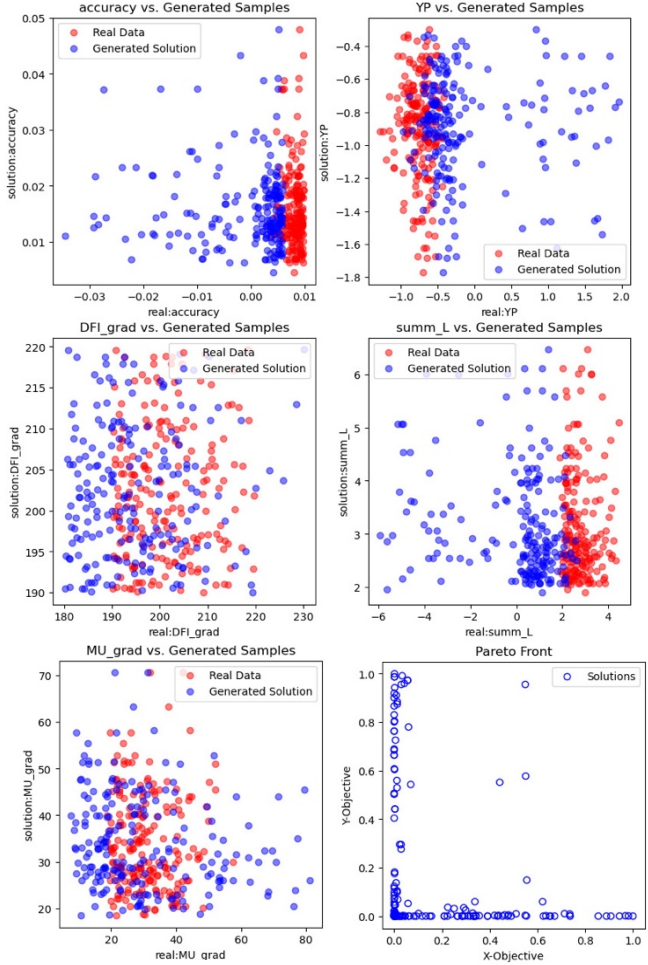

Fig. 7. Obtained results

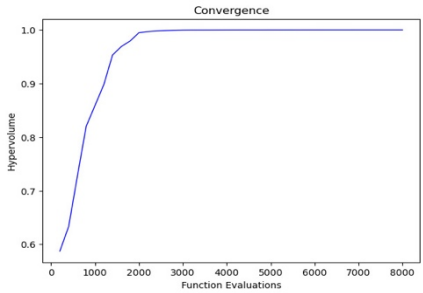

Fig. 8. Convergence of the applied NSGA-II algorithm

Fig.8 unambiguously reveal that the algorithm successfully navigates towards a convergence hypervolume of 1.0, a feat achieved approximately within 2000 iterations.

## V. OPTIMIZATION BASED ON ISOTROPY CRITERION

Consider a manipulator with $W_q = n$ DoF and generalized coordinates $\vec{q} = [q_1, q_2, \ldots, q_n]^T$. Assume that the output object also has $W_x = n$ DoF. Then the singular positions of the manipulator, are determined by the Jacobian matrix $J$: $\det J = 0$. These positions define the boundaries of the working area of the manipulator and are worst in terms of force transfer. The same Jacobian matrix is used in [24] to define the best configurations in terms of force and motion transfer:

$$J^T J = \alpha^2 E \quad \text{or} \quad JJ^T = \alpha^2 E, \quad (11)$$

where E is the identity matrix and $\alpha$ is some real number, the configurations satisfying this criterion are called "isotropic", which are the furthest configurations from singularity. Authors of this study proposed for the first time the application of this criterion on the WR.

Consider a tripod gait, when three legs are in the support, three are always in the transfe phase. In the equivalent scheme (Fig.9) the first, third and fifth legs are in the support phases (feet $S_1, S_3, S_5$), $C$ is the center of mass of the robot body, $O_0\xi\eta\zeta$ is a global coordinate system fixed with the bearing surface, $CXY$ is a coordinate system fixed with the robot body/hull. $O_iP_i$ $(i = 1, 3, 5)$ is a local coordinate system, fixed with a link $O_iP_i$. The local coordinates of the joint $S_i$ in this coordinate system are $x_{S_i} = a_i, y_{S_i} = q_i$, where $q_i$ are generalized coordinates, $i = 1, 3, 5$.

Jacobian matrix $J_q$ of the system which is defined as $\dot{\vec{x}} = J_q\dot{\vec{q}}$, or $\vec{F}_q = -J_q^T\vec{F}_x$, from where $J_q = \frac{d\vec{x}}{d\vec{q}}$, $\vec{F}_q$ is the generalized input forces, $\vec{F}_x$ is a vector $[\vec{F}, \vec{M}]$, $\vec{F}$ and $\vec{M}$ are main output force and momentum.

Let us define the output parameters of the system as $\vec{x} = \left[\vec{R_C}^T, L_\theta\theta\right]^T = [\xi_C, \eta_C, L_\theta\theta]^T$, where $L_\theta$ – characteristic length, and the inputs as $\vec{q} = [q_1, q_3, q_5]^T$.

$J_q$ can be found from kinematic analysis of the system as follows:

$$J_q = A^{-1}B, \quad (12)$$

where

$$A = \begin{bmatrix} \vec{r_{S_1}}^T \cdot \Gamma^T(\theta + \alpha_1) & \frac{1}{L_\theta}\vec{r_{S_1}}^T \cdot \Gamma^T\left(\frac{\pi}{2} - \alpha_1\right) \cdot \vec{r_{O_1}} \\ \vec{r_{S_3}}^T \cdot \Gamma^T(\theta + \alpha_3) & \frac{1}{L_\theta}\vec{r_{S_3}}^T \cdot \Gamma^T\left(\frac{\pi}{2} - \alpha_3\right) \cdot \vec{r_{O_3}} \\ \vec{r_{S_5}}^T \cdot \Gamma^T(\theta + \alpha_5) & \frac{1}{L_\theta}\vec{r_{S_5}}^T \cdot \Gamma^T\left(\frac{\pi}{2} - \alpha_5\right) \cdot \vec{r_{O_5}} \end{bmatrix},$$

$$B = -\begin{bmatrix} q_1 & 0 & 0 \\ 0 & q_3 & 0 \\ 0 & 0 & q_5 \end{bmatrix}.$$

The isotropy condition (12) can be transformed to another form [24]:

$$\left(J_q^{-1}\right)^T \cdot J_q^{-1} = \frac{1}{\lambda^2}E, \quad (13)$$

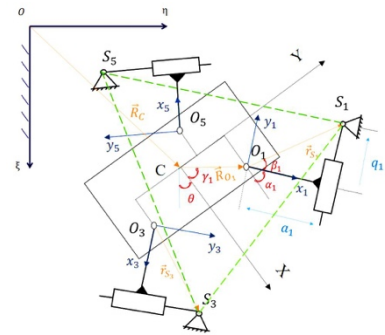

Fig. 9. The equivalent KS of the WR during "tripod"-gate

$$J_q^{-1} =$$

$$\begin{bmatrix} \frac{1}{q_1}\vec{r}_{S_1}^T \cdot \Gamma^T(\theta + \alpha_1) & \frac{1}{q_1 L_\theta}\vec{r}_{S_1}^T \cdot \Gamma^T\left(\frac{\pi}{2} - \alpha_1\right)\cdot \vec{r}_{O_1} \\ \frac{1}{q_3}\vec{r}_{S_3}^T \cdot \Gamma^T(\theta + \alpha_3) & \frac{1}{q_3 L_\theta}\vec{r}_{S_3}^T \cdot \Gamma^T\left(\frac{\pi}{2} - \alpha_3\right)\cdot \vec{r}_{O_3} \\ \frac{1}{q_5}\vec{r}_{S_5}^T \cdot \Gamma^T(\theta + \alpha_5) & \frac{1}{q_5 L_\theta}\vec{r}_{S_5}^T \cdot \Gamma^T\left(\frac{\pi}{2} - \alpha_5\right)\cdot \vec{r}_{O_5} \end{bmatrix} \quad (14)$$

$J_q^{-1}$ can be transformed:

$$J_q^{-1} = \quad (15)$$

$$-\begin{bmatrix} \frac{1}{q_1}\vec{e}_\xi^T \Gamma(\theta + \alpha_1)\vec{r}_{S_1} & \frac{1}{q_1}\vec{e}_\eta^T \Gamma(\theta + \alpha_1)\vec{r}_{S_1} & \frac{1}{q_1 L_\theta}\vec{e}_\zeta^T[\vec{r}_{O_1} \times \Gamma(\alpha_1)\vec{r}_{S_1}] \\ \frac{1}{q_3}\vec{e}_\xi^T \Gamma(\theta + \alpha_3)\vec{r}_{S_3} & \frac{1}{q_3}\vec{e}_\eta^T \Gamma(\theta + \alpha_3)\vec{r}_{S_3} & \frac{1}{q_3 L_\theta}\vec{e}_\zeta^T[\vec{r}_{O_3} \times \Gamma(\alpha_3)\vec{r}_{S_3}] \\ \frac{1}{q_5}\vec{e}_\xi^T \Gamma(\theta + \alpha_5)\vec{r}_{S_5} & \frac{1}{q_5}\vec{e}_\eta^T \Gamma(\theta + \alpha_5)\vec{r}_{S_5} & \frac{1}{q_5 L_\theta}\vec{e}_\zeta^T[\vec{r}_{O_5} \times \Gamma(\alpha_5)\vec{r}_{S_5}] \end{bmatrix}$$

The isotropy conditions are defined as follows:

$$\sum_{i=1,3,5} \frac{\cos^2(\theta + \alpha_i + \beta_i)}{\sin^2 \beta_i} = \frac{1}{\lambda^2}; \quad (16)$$

$$\sum_{i=1,3,5} \frac{\sin^2(\theta + \alpha_i + \beta_i)}{\sin^2 \beta_i} = \frac{1}{\lambda^2}; \quad (17)$$

$$\sum_{i=1,3,5} \frac{r_{O_i} \sin^2(\alpha_i + \beta_i - \gamma_i)}{\sin^2 \beta_i} = \frac{1}{\lambda^2}; \quad (18)$$

$$\sum_{i=1,3,5} \frac{\sin 2(\theta + \alpha_i + \beta_i)}{\sin^2 \beta_i} = 0; \quad (19)$$

$$\sum_{i=1,3,5} \frac{r_{O_i} \cos(\theta + \alpha_i + \beta_i)\sin(\alpha_i + \beta_i - \gamma_i)}{\sin^2 \beta_i} = 0; \quad (20)$$

$$\sum_{i=1,3,5} \frac{r_{O_i} \sin(\theta + \alpha_i + \beta_i)\sin(\alpha_i + \beta_i - \gamma_i)}{\sin^2 \beta_i} = 0. \quad (21)$$

After conducting case studies with various solutions that satisfy the aforementioned isotropy conditions, the following solution is selected as optimal [18]:

$u_1 = u_3 = u_5$, where $u_i = r_{O_i}\sin(\alpha_i + \beta_i - \gamma_i)$, $i = 1,3,5$,

1. $\alpha_3 = \alpha_1 - \frac{2\pi}{3}, \alpha_5 = \alpha_1 + \frac{2\pi}{3}$,

2. $\alpha_3 = \alpha_1 + \frac{2\pi}{3}, \alpha_5 = \alpha_1 - \frac{2\pi}{3}$

$$r_{O_1} = r_{O_3} = r_{O_5} = \pm \frac{L_\theta}{\sqrt{2}\sin(\alpha_1 + \beta - \gamma_1)};$$

$$\gamma_3 - \gamma_1 = \alpha_3 - \alpha_1;$$

$$\gamma_5 - \gamma_1 = \alpha_5 - \alpha_1.$$

For $\gamma_1 = \frac{\pi}{3}$,

1) for the first solution, $\gamma_3 = \gamma_1 - \frac{2\pi}{3} = -\frac{\pi}{3}; \gamma_5 = \gamma_1 + \frac{2\pi}{3} = \pi$.

2) and for the second solution $\gamma_3 = \gamma_1 + \frac{2\pi}{3} = \pi; \gamma_5 = \gamma_1 - \frac{2\pi}{3} = -\frac{\pi}{3}$.

Note that the second configuration can be obtained from the first by swapping leg mechanisms with numbers 3 and 5.

And for $\gamma_1 = -\frac{\pi}{3}$,

1) the first solution is $\gamma_3 = -\pi$; $\gamma_5 = \frac{\pi}{3}$,

2) and the second is $\gamma_3 = \frac{\pi}{3}$; $\gamma_5 = -\pi$,

i.e. swapped are the legs with numbers 1 and 3.

## VI. EXPERIMENTAL PROTOTYPE WITH ADAPTATION TO IRREGULARITIES AND SLAM NAVIGATION FOR A ROBOT

### A. Adaptation to Surface Irregularities and Laboratory Prototypes

In this section an adaptation mechanism is introduced that ensures smooth horizontal movement of the robot body by *eliminating vertical displacement* on uneven surfaces, and has been tested on the physical prototype. Future developments aim to incorporate a locking device within the adaptation system, ensuring that the *weight of the WR body is supported by this locking device rather than the motors*, thereby substantially reducing energy usage. The test of the adaptation system is demonstrated in Fig.10. During the transfer phase the adaptation motor raises the foot to the maximum height. At the rotation angle $\varphi_1 = \varphi_0 + 360°$, the leg enters the support phase, the foot begins to lower, and upon contact with the support surface, the foot contact sensor sends a signal to the adaptation engine to "lock the engine".

Multiple iterations of experimental laboratory models of the WR were constructed (Fig.11), testing the functionality of each mechanism of the robot one-by-one. It was demonstrated the complete functionality of the design and affirmed the validity of the primary hypotheses.

### B. SLAM Algorithm

The SLAM process can be mathematically described by the following steps:

Initialization: The robot starts with an initial position estimate and an empty map. Let $x_0$ be the initial position estimate and $M_0$ be the initial map.

Prediction: As the robot moves, its odometry data provides an estimate of its new position. If $u_t$ represents the control input at time $t$, $f$ is the motion, model the predicted state $x_t'$:

$$x_t' = f(x_{t-1}, u_t). \quad (22)$$

Observation: The robot uses its sensors (e.g., 3D Lidar) to observe the environment, generating a set of measurements $z_t$. These measurements are used to update the map and the robot's position. The observation model is represented as:

$$z_t = h(x_t, M_t) + w_t, \quad (23)$$

$h$ is the observation function and $w_t$ is the observation noise.

Correction: The robot updates its position $x_t$ and map based on the new measurements. The updated state and map $M_t$ are given by:

$$x_t = x_t' + K_t(z_t - h(x_t', M_t)) \quad (24)$$

$$M_t = M_{t-1} + \Delta M_t \quad (25)$$

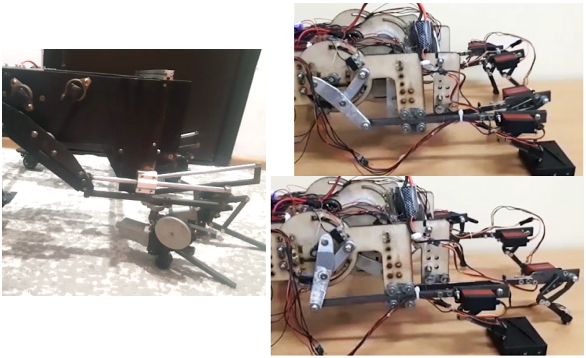

Fig. 10. Testing the adaptation mechanism of the WR

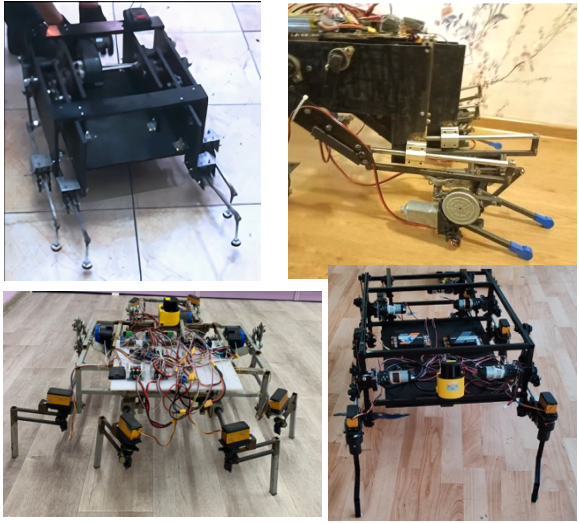

Fig. 11. WR prototypes

The process of integrating 3D Lidar data into SLAM involves:

Point Cloud Generation: The Lidar sensor generates a point cloud, which is a collection of data points in space representing the environment.

Data Fusion: The point cloud data is fused with odometry data to correct any drifts and improve localization accuracy. $\hat{x}_t$ is the corrected state estimate. Then the fusion process typically uses a Kalman filter, represented by:

$$\hat{x}_t = x'_t + K_t(z_t - h(x'_t, M_t)) \qquad (26)$$

Map Update: The corrected position and point cloud data are used to update the map.

Fig.12 illustrates the path planning process utilizing SLAM with a 3D Lidar sensor. The 3D Lidar provides high-resolution point cloud data, which the SLAM algorithm uses to dynamically map the environment and plan optimal paths. This integration ensures precise navigation by continuously updating the map and localizing the robot within complex environments.

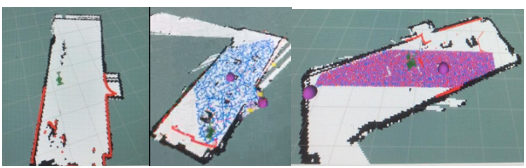

Fig. 12. Path planning

ACKNOWLEDGEMENTS

The authors would like to express their sincere gratitude for the financial support provided by the Fundamental Research Grant from the Ministry of Science and Higher Education of the Republic of Kazakhstan (Grant Number: BR20280990).